\documentclass{article}

\usepackage{arxiv}

\usepackage[utf8]{inputenc} 
\usepackage[T1]{fontenc}    
\usepackage{hyperref}       
\usepackage{url}            
\usepackage{booktabs}       
\usepackage{amsfonts}       
\usepackage{nicefrac}       
\usepackage{microtype}      
\usepackage{lipsum}

\usepackage{graphicx}
\usepackage{microtype}
\usepackage{hyperref}
\usepackage[numbers,sort&compress]{natbib}

\usepackage{soul}
\usepackage{graphicx}
\usepackage[utf8]{inputenc} 
\usepackage[T1]{fontenc}    
\usepackage{hyperref}       
\usepackage{url}            
\usepackage{booktabs}       
\usepackage{amsfonts}       
\usepackage{nicefrac}       
\usepackage{microtype}      
\usepackage{amsmath}
\usepackage{diagbox}
\usepackage{amssymb}
\usepackage{multirow}
\usepackage{comment}        
\usepackage[normalem]{ulem} 
\usepackage{slashbox}
\usepackage{caption}
\usepackage{subcaption}
\usepackage{mathtools}
\newbox{\bigpicturebox}

\usepackage{amsmath,amsfonts,bm}

\def\eqref#1{equation~\ref{#1}}

\def\1{\bm{1}}

\def\va{{\bm{a}}}

\def\vh{{\bm{h}}}

\def\vm{{\bm{m}}}

\def\vx{{\bm{x}}}

\def\eva{{a}}

\def\mA{{\bm{A}}}

\def\mD{{\bm{D}}}

\def\mH{{\bm{H}}}
\def\mI{{\bm{I}}}

\def\mL{{\bm{L}}}
\def\mM{{\bm{M}}}

\def\mP{{\bm{P}}}

\def\mU{{\bm{U}}}
\def\mV{{\bm{V}}}
\def\mW{{\bm{W}}}
\def\mX{{\bm{X}}}

\DeclareMathAlphabet{\mathsfit}{\encodingdefault}{\sfdefault}{m}{sl}
\SetMathAlphabet{\mathsfit}{bold}{\encodingdefault}{\sfdefault}{bx}{n}

\def\gG{{\mathcal{G}}}

\def\sN{{\mathcal{N}}}

\def\sR{{\mathbb{R}}}

\def\emA{{A}}

\def\emD{{D}}

\newtheorem{remark}{Remark}
\renewcommand{\vec}[1]{\boldsymbol{#1}}

\newcommand*{\model}{ASGAT}
\newcommand*{\citeseer}{{\sc CiteSeer}}
\newcommand*{\cora}{{\sc Cora}}
\newcommand*{\pubmed}{{\sc Pubmed}}
\newcommand*{\cornell}{{\sc Cornell}}
\newcommand*{\texas}{{\sc Texas}}
\newcommand*{\wisconsin}{{\sc Wisconsin}}
\newcommand*{\chameleon}{{\sc Chameleon}}
\newcommand*{\squirrel}{{\sc Squirrel}}

\title{Beyond Low-Pass Filters: Adaptive Feature Propagation on Graphs}

\author{Shouheng Li \\
The Australian National University \\
Canberra, Australia \\
shouheng.li@anu.edu.au
\And
Dongwoo Kim \\
Pohang University of Science and Technology \\
Pohang, South Korea\\

\And
Qing Wang \\
The Australian National University \\
Canberra, Australia}

\date{}
\begin{document}
\maketitle
\let\thefootnote\relax\footnotetext{ECML-PKDD 2021, Bilbao, Spain}
\begin{abstract}
Graph neural networks (GNNs) have been extensively studied for prediction tasks on graphs. As pointed out by recent studies, most GNNs assume local homophily, i.e., strong similarities in local neighborhoods. This assumption however limits the generalizability power of GNNs. 
To address this limitation, we propose a flexible GNN model, which is capable of handling any graphs without being restricted by their underlying homophily. 
At its core, this model adopts a node attention mechanism based on multiple learnable spectral filters; therefore, the aggregation scheme is learned adaptively for each graph in the spectral domain.
We evaluated the proposed model on node classification tasks over eight benchmark datasets. The proposed model is shown to generalize well to both homophilic and heterophilic graphs. Further, it outperforms all state-of-the-art baselines on heterophilic graphs and performs comparably with them on homophilic graphs. 

\end{abstract}

\section{Introduction}

Graph neural networks (GNNs) have recently demonstrated great power in graph-related learning tasks, such as node classification~\citep{DBLP:conf/iclr/KipfW17}, link prediction~\citep{NIPS2018_7763} and graph classification \citep{DBLP:conf/kdd/LeeRK18}. Most GNNs follow a message-passing architecture where, in each GNN layer, a node aggregates information from its direct neighbors indifferently. In this architecture, information from long-distance nodes is propagated and aggregated by stacking multiple GNN layers together~\citep{DBLP:conf/iclr/KipfW17, DBLP:conf/iclr/VelickovicCCRLB18, DBLP:conf/nips/DefferrardBV16}. However, this architecture underlies the assumption of local homophily, i.e. proximity of similar nodes.
While this assumption seems reasonable and helpful to achieve good prediction results on homophilic graphs such as citation networks~\citep{Pei2020}, it limits GNNs' generalizability to heterophilic graphs. 
Heterophilic graphs commonly exist in the real-world, for instance, people tend to connect to opposite gender in dating networks, and different amino acid types are more likely to form connections in protein structures \citep{DBLP:conf/nips/ZhuYZHAK20}. Moreover, determining whether a graph is homophilic or not is a challenge by itself. In fact, strong and weak homophily can both exhibit in different parts of a graph, which makes a learning task more challenging. 



\citet{Pei2020} proposed a metric to measure local node homophily based on how many neighbors of a node are from the same class. Using this metric, they categorized graphs as homophilic (strong homophily) or heterophilic (weak homophily), and showed that classical GNNs such as GCN~\citep{DBLP:conf/iclr/KipfW17} and GAT~\citep{DBLP:conf/iclr/VelickovicCCRLB18} perform poorly on heterophilic graphs. \citet{Liu2020} further showed that GCN and GAT are outperformed by a simple multi-layer perceptron (MLP) in node classification tasks on heterophilic graphs. This is because the naive local aggregation of homophilic models brings in more noise than useful information for such graphs. These findings indicate that these GNN models perform sub-optimally when the fundamental assumption of homophily does not hold.

Based on the above observation, we argue that a well-generalized GNN should perform well on graphs regardless of homophily. Furthermore, since a real-world graph can exhibit both strong and weak homophily in different node neighborhoods, a powerful GNN model should be able to aggregate node features using different strategies accordingly. For instance, in heterophilic graphs where a node shares no similarity with any of its direct neighbors, such a GNN model should be able to ignore direct neighbors and reach farther to find similar nodes, or at least, resort to the node's attributes to make a prediction. Since the validity of the assumption about homophily is often unknown, such aggregation strategies should be learned from data rather than decided upfront. 

To circumvent this issue, in this paper, we propose a novel GNN model with attention-based adaptive aggregation, called \model. Most existing attention-based aggregation architectures perform self-attention to the local neighborhood of a node~\citep{DBLP:conf/iclr/VelickovicCCRLB18}. Unlike these approaches, we aim to design an aggregation method that can gather informative features from both close and far-distant nodes. To achieve this, we employ graph wavelets under a relaxed condition of localization, which enables us to learn attention weights for nodes in the spectral domain. In doing so, the model can effectively capture information from frequency components and thus aggregate both local information and global structure into node representations.

To further improve the generalizability of our model, instead of using predefined spectral kernels, we propose to use multi-layer perceptrons (MLP) to learn desired spectral filters without limiting their shapes. Existing works on graph wavelet transform choose wavelet filters heuristically, such as heat kernel, wave kernel and personalized page rank kernel \citep{DBLP:conf/nips/KlicperaWG19,DBLP:conf/iclr/XuSCQC19,DBLP:conf/iclr/KlicperaBG19}. They are mostly low-pass filters, which means that these models implicitly treat high-frequency components as ``noises'' and have them discarded~\citep{DBLP:journals/corr/abs-1905-09550, DBLP:journals/spm/ShumanNFOV13, Hammond2011, DBLP:journals/corr/abs-2003-07450}. However, this may hinder the generalizability of models since high-frequency components can carry meaningful information about local discontinuities, as analyzed in \citep{DBLP:journals/spm/ShumanNFOV13}. Our model overcomes these limitations using node attentions derived from fully learnable spectral filters.

To summarize, the main contributions of this work are as follows:
\begin{enumerate}
  \item We show that high-frequency components carry important information on heterophilic graphs which can be used to improve prediction performance.
  \item We propose a generalized GNN model which performs well on both homophilic and heterophilic graphs, regardless of graph homophily.
  \item We exhibit that multi-headed attention produced by multiple spectral filters work better than attention obtained from a single filter, as it enables flexibility to aggregate features from different frequency components.
\end{enumerate}


We conduct extensive experiments to compare \model{} with well-known baselines on node classification tasks. The experimental results show that \model{} significantly outperforms the state-of-the-art methods on heterophilic graphs where local node homophily is weak, and performs comparably with the state-of-the-art methods on homophilic graphs where local node homophily is strong. This empirically verifies that \model{} is a general model for learning on different types of graphs.\footnote{Our open-sourced code is available at \url{https://github.com/seanli3/asgat}.}
\section{Preliminaries}





Let $\gG=(V,E,A, \boldsymbol{x})$ be an undirected graph with $N$ nodes, where $V$, $E$, and $A$ are the node set, edge set, and adjacency matrix of $\gG$, respectively, and $\boldsymbol{x}: V\mapsto \mathbb{R}^m$ is a graph signal function that associates each node with a feature vector. The normalized Laplacian matrix of $\gG$ is defined as $\mL = \mI - \mD^{-1/2}\mA\mD^{-1/2}$, where $\mD \in {\mathbb R}^{N\times N}$ is the diagonal degree matrix of $\gG$. 
In spectral graph theory, the eigenvalues $\Lambda = \text{diag}(\lambda_1,...,\lambda_N)$ and eigenvectors $\mU$ of $\mL = \mU\Lambda \mU^H$ are known as the graph's spectrum and spectral basis, respectively, where $\mU^H$ is the Hermitian transpose of $\mU$. The graph Fourier transform of $\boldsymbol{x}$ is $\boldsymbol{\hat{x}} = \mU^H\boldsymbol{x}$ and its inverse is $\boldsymbol{x} = \mU\hat{\boldsymbol{x}}$.

The spectrum and spectral basis carry important information on the connectivity of a graph~\citep{DBLP:journals/spm/ShumanNFOV13}. Intuitively, lower frequencies correspond to global and smooth information on the graph, while higher frequencies correspond to local information, discontinuities and possible noise~\citep{DBLP:journals/spm/ShumanNFOV13}. One can apply a spectral filter and use graph Fourier transform to manipulate signals on a graph in various ways, such as smoothing and denoising~\citep{DBLP:conf/globalsip/SchaubS18}, abnormally detection~\citep{5967745} and clustering~\citep{8462239}.
Spectral convolution on graphs is defined as the multiplication of a signal $\vx$ with a filter $g(\Lambda)$ in the Fourier domain, i.e.
\begin{equation}\label{eqn:spectral_filter}
g(\mL)\vec{x} = g(\mU\Lambda \mU^H)\vec{x} = \mU g(\Lambda)\mU^H\vec{x} = \mU g(\Lambda)\hat{\vec{x}}.
\end{equation}
When a spectral filter is parameterized by a scale factor, which controls the radius of neighbourhood aggregation, \autoref{eqn:spectral_filter} is also known as the Spectral Graph Wavelet Transform (SGWT)~\citep{Hammond2011, DBLP:journals/spm/ShumanNFOV13}. For example, \citet{DBLP:conf/iclr/XuSCQC19} uses a small scale parameter $s < 2$ for a heat kernel, $g(s\lambda) = e^{-\lambda s}$, to localize the wavelet at a node. 

\section{Proposed Approach}

\begin{figure*}[!t]
\centering
\includegraphics[width=0.95\textwidth]{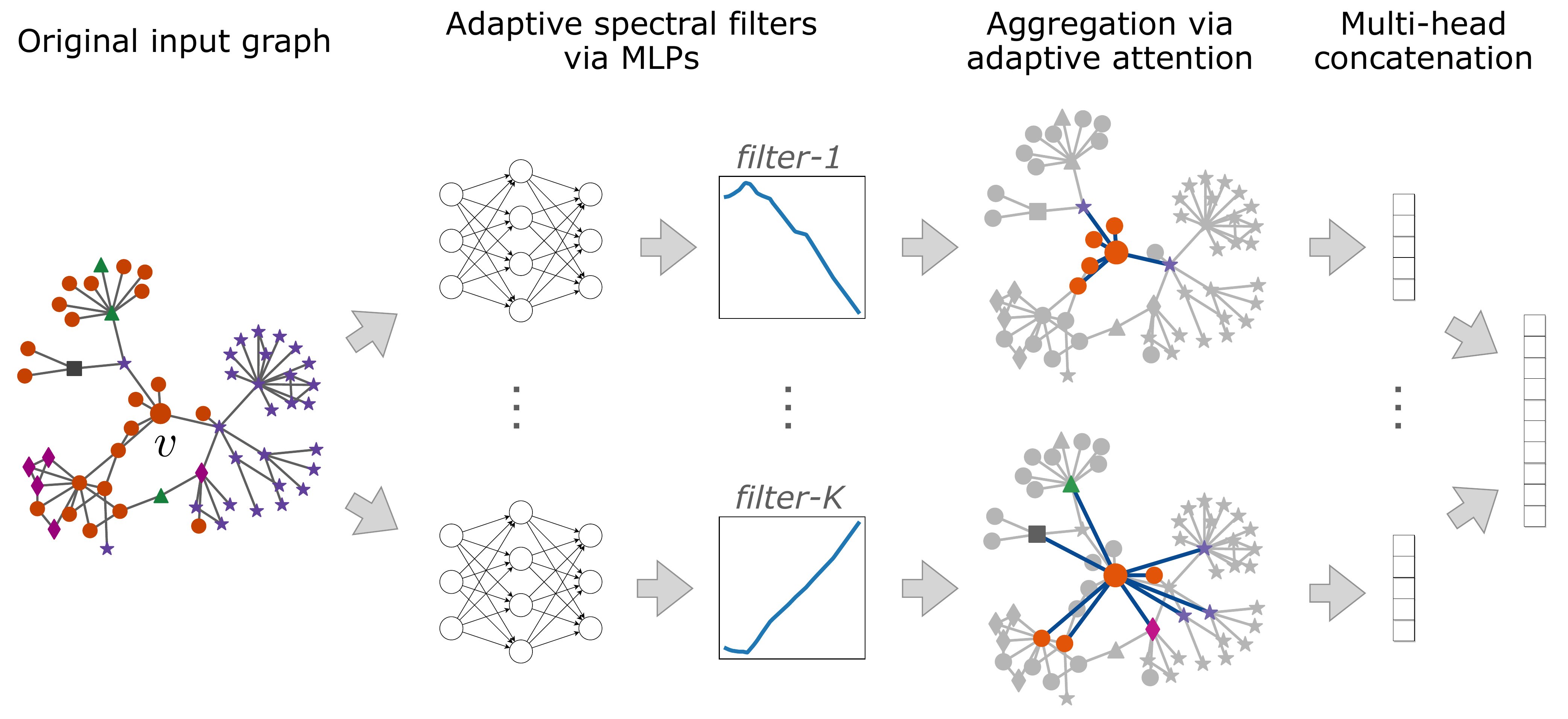}
\caption{Illustration of a spectral node attention layer on a three-hop ego network of the central node $v$ from the \citeseer{} dataset. Node classes are indicated by shape and color. Passing the graph through two learned spectral filters place attention scores on nodes, including node $v$ itself. Nodes with positive attention scores are presented in color. Node features are aggregated for node $v$ according to attention scores. The low-pass filter attend to local neighbors (filter 1), while the high-pass filter skips the first hop and attend the nodes in the second hop (filter $K$). The resulting embeddings from multiple heads are then concatenated before being sent to the next layer (multi-head concatenation). Note that we have visualized learned filters from experiments.}
\label{fig:illustration}
\end{figure*}

Graph neural networks (GNNs) learn lower-dimensional embeddings of nodes from graph structured data. In general, given a node, GNNs iteratively aggregate information from its neighbor nodes, and then combine the aggregated information with its own information. An embedding of node $v$ at the $l$th layer of GNN is typically formulated as
\begin{align*}
\vm_v &= \text{aggregate}(\{\vh^{(l-1)}_u | u \in \sN_v \}) \\
\vh^{(l)}_v &= \text{combine}(\vh^{(l-1)}_v, \vm_v),
\end{align*}
where $\sN_v$ is the set of neighbor nodes of node $v$, $\vm_v$ is the aggregated information from the neighbors, and $\vh^{(l)}_v$ is the embedding of node $v$ at the $l$th layer ($\vh^{(0)}_v = \vx_v$). The embedding $\vh^{(L)}_v$ of node $v$ at the final layer is then used for some prediction tasks. In most GNNs, $\sN_v$ is restricted to a set of one-hop neighbors of node $v$. Therefore, one needs to stack multiple aggregation layers in order to collect the information from more than one-hop neighborhood within this architecture.

\textbf{Adaptive spectral filters.} Instead of stacking multiple aggregation layers, we introduce a spectral attention layer that rewires a graph based on spectral graph wavelets. A spectral graph wavelet $\boldsymbol\psi_v$ at node $v$ is a modulation in the spectral domain of signals centered around the node $v$, given by an $N$-dimensional vector
\begin{align}
\boldsymbol\psi_v = \mU g(\Lambda) \mU^{H} \delta_v,
\label{eqn:wavelet}
\end{align}
where $g(\cdot)$ is a spectral filter and $\delta_v$ is a one-hot vector for node $v$.

The common choice of a spectral filter is a heat kernel. A wavelet coefficient $\psi_{vu}$ computed from a heat kernel can be interpreted as the amount of energy that node $v$ has received from node $u$ in its local neighborhood. In this work, instead of using pre-defined localized kernels, we use multi-layer perceptrons (MLP) to learn spectral filters. With learnable spectral kernels, we obtain the inverse graph wavelet transform
\begin{align}\label{eqn:graph_wavelet_matrix}
\boldsymbol\psi_v = \mU \text{diag}(\text{MLP}(\Lambda)) \mU^H \delta_v.
\end{align}

\noindent Unlike a low-pass heat kernel, where the wavelet coefficients can be understood as the amount of energy after heat diffusion, the learned coeffcients $\psi_{vu}$ do not always correspond to energy diffusion. In spectral imaging processing, lower frequency components preserve an image's background, while higher frequency components are useful to detect object edges or outlines. Similarly, in spectral graph theory, lower-frequency components carry smoothly changing signals. Therefore a low-pass filter is a reasonable choice to extract features and denoise a homophilic graph. In contrary, higher-frequency components carry abruptly changing signals, which correspond to the discontinuities and "opposite attraction" characteristics of heterophilic graphs. Since $\text{MLP}(\Lambda)$ is learned in training, it can learn a low-pass filter which works as a diffusion operator, while in other cases, especially on heterophilic graphs, it converges to a high-pass filer at most times (\autoref{sec:experiments}).

 Note that we use the terminology wavelet and spectral filter interchangeably as we have relaxed the wavelet definition from \citep{Hammond2011} so that learnable spectral filters in our work are not necessarily localized in the spectral and spatial domains.
  \begin{remark}
 \autoref{eqn:graph_wavelet_matrix} requires the eigen-decomposition of a Laplacian matrix, which is expensive and infeasible for large graphs. To address this computational issue, one may use well-studied methods such as Chebyshev~\citep{Hammond2011, DBLP:conf/iclr/XuSCQC19, DBLP:conf/nips/KlicperaWG19} and Auto-Regressive Moving-Average (ARMA)~\citep{Isufi2017, Liu2019} to efficiently compute an approximate the graph filtering of MLP in \autoref{eqn:graph_wavelet_matrix}.
 \end{remark}

\textbf{Attention mechanism.~}
Unlike the previous work~\citep{DBLP:conf/iclr/XuSCQC19} where the output of inverse graph wavelet transform are directly used to compute node embeddings, we normalize the output through a softmax layer
\begin{align}
\label{eqn:softmax}
\va_v = \text{softmax}(\boldsymbol\psi_v),
\end{align}

\noindent where $\va_v\in \mathbb{R}^N$ is an attention weight vector. With attention weights, an update layer is then formalized as
\begin{align}
\label{eqn:layer}
\vh^{(l)}_v =  \sigma\left(\sum_{u=1}^{N} \eva_{vu} \vh^{(l-1)}_u \mW^{(l)}\right),    
\end{align}
where $\mW^{(l)}$ is a weight matrix shared across all nodes at the $l$th layer and $\sigma$ is ELU nonlinear activation.  

Note that the update layer is not divided into aggregation and combine steps in our work. Instead, we compute the attention $\eva_{vv}$ directly from a spectral filter. Unlike heat kernel and other spectral filters, the output of inverse graph wavelet transform with a learnable spectral kernel are not always localized. Hence, the model can adaptively aggregate information from both close and far-distant nodes, depending on their attention weights.

\textbf{Sparsified node attentions.} 
With predefined localized spectral filters such as a heat kernel, most of wavelet coefficients are zero due to their locality. In our work, spectral filters are fully learned from data, consequently attention weights obtained from learnable spectral filters do not impose any sparsity. This means that to perform an aggregation operation we need to retrieve all possible nodes in a graph, which is not efficient. From our experiments, we observe that most attention weights are negligible after softmax. 
Thus, we consider a sparsification technique to keep only the largest $k$ entries of \autoref{eqn:graph_wavelet_matrix} for each node, i.e.
  \begin{align}
        \label{eqn:psi_k}
        \bar{{\psi}}_{vu} = 
        \begin{cases} 
            {\psi}_{vu} & \quad \text{ if } {\psi}_{vu} \in \text{topK}(\{\psi_{v0}, ..., \psi_{vN}\}, k) \\
            -\infty & \quad \text{ otherwise},
        \end{cases}
    \end{align}
where topK is a partial sorting function that returns the largest $k$ entries from a set of wavelet bases $\{\psi_{v0}, ..., \psi_{vN}\}$. This technique guarantees attention sparsity such that the embedding of each node can be aggregated from at most $k$ other nodes with a time complexity trade-off of $O(N+k\log N)$.
The resulting $\bar{\psi}$ is then fed into the softmax layer to compute attention weights.

We adopt multi-head attention to model multiple spectral filters. Each attention head aggregates node information with a different spectral filter, and the aggregated embedding is concatenated before sent to the next layer. 
To reduce redundancy, we adopt a single MLP$:\sR^N \rightarrow \sR^{N\times M}$, where $M$ is the number of attention heads, and each column of the output corresponds to one adaptive spectral filter.

We name the multi-head spectral attention architecture as a \emph{adaptive spectral graph attention network} (\model{}). The design of \model{} is easily generalizable, and many existing GNNs can be expressed as special cases of \model{} (see \autoref{sec:connection}). \autoref{fig:illustration} illustrates how \model{} works with two attention heads learned from the \citeseer{} dataset. As shown in the illustration, the MLP learns adaptive filters such as low-pass and high-pass filters. A low-pass filter assigns high attention weights in local neighborhoods, while a high-pass filter assigns high attention weights on far-distant but similar nodes, which cannot be captured by a traditional hop-by-hop aggregation scheme.



\section{Experiments}
\label{sec:experiments}

To evaluate the performance of our proposed model, we conduct experiments on node classification tasks with homophilic graph datasets, and heterophilic graph datasets. Further ablation study highlights the importance of considering the entire spectral frequency.




\subsection{Experimental Setup}

\begin{table*}[t!]
     \centering
     \caption{Micro-F1 results for node classification. The proposed model consistently outperforms the GNN methods on heterophilic graphs and performs comparably on homophilic graphs. Results marked with $\dagger$ are obtained from \citet{Pei2020}. Results marked with $\ddag$ are obtained from \citet{DBLP:conf/nips/ZhuYZHAK20}.}
     \scalebox{0.77}{
         \begin{tabular}{l | r  r  r | r  r  r  r r} 
             \toprule
              & \multicolumn{8}{c}{$\text{Homophily} \xLeftrightarrow{\hspace*{6cm}} \text{Heterophily}$} \\
              & \multicolumn{1}{c}{\cora} & \multicolumn{1}{c}{\pubmed} & \multicolumn{1}{c|}{\citeseer} & \multicolumn{1}{c}{\chameleon} &
              \multicolumn{1}{c}{\squirrel} & \multicolumn{1}{c}{\wisconsin} & \multicolumn{1}{c}{\cornell} & \multicolumn{1}{c}{\texas} \\              
              \midrule
              \midrule
              $\beta$ & 0.83 & 0.79 & 0.71 & 0.25 & 0.22 & 0.16 & 0.11 & 0.06\\
              \#Nodes & 2,708 & 19,717 & 3,327 & 2,277 & 5,201 & 251 & 183 & 183 \\
              \#Edges & 5,429 & 44,338 & 4,732 & 36,101 & 217,073 & 515 & 298 & 325\\
              \#Features & 1,433 & 500 & 3,703 & 2,325 & 2,089 & 1,703 & 1,703 & 1,703 \\
              \#Classes & 7 & 3 & 6 & 5 & 5 & 5 & 5 & 5\\
             \midrule
             GCN & $87.4\pm 0.2$ & $87.8\pm0.2$ & $78.5\pm0.5$ & ${59.8\pm2.6}^\ddag$ & ${36.9\pm1.3}^\ddag$ & $64.1\pm6.3$ & $59.2\pm3.2$& $64.1\pm4.9$ \\
             ChevNet & $88.2\pm 0.2$ & $89.3\pm0.3$ & ${79.4}\pm0.4$  & $66.0\pm2.3$ & $39.6\pm3.0$ & $82.5\pm2.8$ & $76.5\pm9.4$ & $79.7\pm5.0$ \\
             ARMANet & $85.2\pm2.5$ & $86.3\pm5.7$ & $76.7\pm0.5$ & $62.1\pm3.6$ & $47.8\pm3.5$ & $78.4\pm4.6$ & $74.9\pm2.9$ & $82.2\pm5.1$  \\ 
             GAT & $87.6\pm 0.3$ & $83.0\pm0.1$ & $77.7\pm0.3$ & ${54.7\pm2.0}^\ddag$  & ${30.6\pm2.1}^\ddag$ & $62.0\pm5.2$ & $58.9\pm3.3$ & $60.0\pm5.7$ \\
             SGC & $87.2\pm0.3$ & $81.1\pm0.3$ & $78.8\pm0.4$ & $33.7\pm3.5$ & $46.9\pm1.7$& $51.8\pm5.9$& $58.1\pm4.6$ & $58.9\pm6.1$\\
             GraphSAGE & $86.3\pm 0.6$ & $89.2\pm0.5$ & $77.4\pm 0.5$ & $51.1\pm0.5$ & ${41.6\pm0.7}^\ddag$ & $77.6\pm4.6$ & $67.3\pm6.9$ &  $82.7\pm4.8$\\
             APPNP & $\textbf{88.4}\pm0.3$ & $86.0\pm0.3$ & $77.6\pm0.6$ & $45.3\pm1.6$ &$31.0\pm1.6$ & $81.2\pm2.5$ & $70.3\pm9.3$ & $79.5\pm4.6$\\
             Geom-GCN & $86.3\pm0.3$ & $89.1\pm0.1$ & $\textbf{81.4}\pm0.3$ & $60.9^\dagger$ & $38.1^\dagger$ & $64.1^\dagger$ & $60.8^\dagger$ & $67.6^\dagger$ \\
             $\text{H}_2\text{GCN}$ & $88.3\pm0.3$ & $89.1\pm0.4$ & $78.4\pm0.5$ & $59.4\pm2.0$ & $37.9\pm2.0$ & $86.5\pm4.4$ & $82.2\pm6.0$ & $82.7\pm5.7$\\  
             \midrule
             MLP & $72.1\pm1.3$ & $88.6\pm0.2$ & $74.9\pm1.8$ & $45.7\pm2.7$ & $28.1\pm2.0$ & $82.7\pm4.5$ & $81.4\pm6.3$ & $79.2\pm6.1$ \\
             \midrule
             Vanilla \model{} & $-$ & $-$ & $-$ & $-$ & $-$ & $\textbf{86.9}\pm4.2$ & $\textbf{84.6}\pm5.8$ & $82.2\pm3.2$ \\ 
             \model-Cheb & $87.5\pm0.5$ & $\textbf{89.9}\pm0.9$ & $79.3\pm0.6$ & $\textbf{66.5}\pm2.8$ & $\textbf{55.8}\pm3.2$ & $86.3\pm3.7$ & $82.7\pm8.3$ & $\textbf{85.1}\pm5.7$ \\
             \model-ARMA & $87.4\pm1.1$ & $88.3\pm1.0$ & $79.2\pm1.4$ & $65.8\pm2.2$ & $51.4\pm3.2$ & $84.7\pm4.4$ & ${83.2}\pm5.5$ & $79.5\pm7.7$ \\
             \bottomrule
        \end{tabular}
    }
    \label{tab:micro_f1}
\end{table*}

\textbf{Baseline methods.~} An exact computation of \autoref{eqn:graph_wavelet_matrix} requires to compute the eigenvectors of the Laplacian matrix, which is often infeasible due to a large graph size. To overcome this issue, we approximate graph wavelet transform response of $\text{MLP}$ with Chebyshev polynomial, dubbed as \model{}-Cheb, and ARMA rational function, dubbed as \model{}-ARMA. We also report the results from the exact computation of eigenvectors whenever possible, which is dubbed as vanilla \model{}.

We compare all variants against 10 benchmark methods, they are vanilla GCN~\citep{DBLP:conf/iclr/KipfW17} and its simplified version SGC~\citep{DBLP:conf/icml/WuSZFYW19}; two spectral methods: ChevNet~\citep{DBLP:conf/nips/DefferrardBV16} and ARMANet~\citep{DBLP:journals/corr/abs-1901-01343}; the graph attention model GAT~\citep{DBLP:conf/iclr/VelickovicCCRLB18}; APPNP, which also adopts adaptive aggregation~\citep{DBLP:conf/iclr/KlicperaBG19}; the neighbourhood-sampling method GraphSage~\citep{DBLP:conf/nips/HamiltonYL17}; Geom-GCN~\citep{Pei2020} and $\text{H}_2\text{GCN}$~\citep{DBLP:conf/nips/ZhuYZHAK20}, both also target prediction on heterophilic graphs. We also include MLP in the baselines since it performs better than many GNN methods on some heterophilic graphs~\citep{Liu2020}.

\textbf{Datasets.~}We evaluate our model and the baseline methods on node classification tasks over three citation networks: \cora{}, \citeseer{} and \pubmed{}~\citep{DBLP:journals/aim/SenNBGGE08}, three webgraphs from the WebKB dataset\footnote{\url{http://www.cs.cmu.edu/afs/cs.cmu.edu/project/theo-11/www/wwkb/}}: \wisconsin, \texas{} and \cornell, and webgraphs from Wikipedia called \chameleon{} and \squirrel{}~\citep{DBLP:journals/corr/abs-1909-13021}.

To quantify the homophily of graphs, we use the metric $\beta$ introduced by \citet{Pei2020},
\begin{equation}
    \beta = \frac{1}{N}\sum_{v\in V}\beta_v\text{\; and \; }
    \beta_v = \frac{|\{u\in\sN_v|\ell(u)=\ell(v)\}|}{|\sN_v|},
\end{equation}
where $\ell(v)$ refers to the label of node $v$.
$\beta$ measures the degree of homophily of a graph, and $\beta_v$ measures the homophily of node $v$ in the graph. A graph has strong local homophily if $\beta$ is large and vice versa. Details of these datasets are summarized in \autoref{tab:micro_f1}.

\textbf{Hyper-parameter settings.~} For citation networks, we follow the experimental setup for node classification from \citep{DBLP:conf/nips/HamiltonYL17, DBLP:conf/nips/Huang0RH18, DBLP:conf/iclr/ChenMX18} and report the results averaged on 10 runs. For webgraphs, we run each model on the 10 splits provided by \citep{Pei2020} and take the average, where each split uses 60\%, 20\%, and 20\% nodes of each class for training, validation and testing, respectively. The results we report on GCN and GAT are better than \citet{Pei2020} as a result of converting the graphs to undirected before training \footnote{\url{https://openreview.net/forum?id=S1e2agrFvS}}. Geom-GCN uses node embeddings pre-trained from different embedding methods such as Isomap~\citep{Tenenbaum2000}, Poincare~\citep{Nickel2017} and struc2vec~\citep{Ribeiro2017}. We report the best micro-F1 results among all three variants for Geom-GCN.

We use the best-performing hyperparameters specified in the original papers of baseline methods. For hyperparameters not specified in the original papers, we use the parameters from \citet{DBLP:journals/corr/abs-1903-02428}. We report the test accuracy results from epochs with the smallest validation loss and highest validation accuracy. Early termination is adopted for both validation loss and accuracy, thus training is stopped when neither validation loss or accuracy improve for 100 consecutive epochs. For \model{}, we use a two-layer architecture where multi-headed filters are learned using a MLP of 2 hidden layers. Each layer of the MLP consists of a linear function and a ReLU activation. To avoid overfitting, dropout is applied in each \model{} layer on both attention weights and inputs equally. Results for vanilla \model{} are only reported for small datasets where eigen-decomposition is feasible.

\subsection{Results and Discussion}
\label{subsec:results}

We use two evaluation metrics to evaluate the performance of node classification tasks: micro-F1 and macro-F1. The results with micro-F1 are summarized in \autoref{tab:micro_f1}. Overall, on homophilic citation networks, \model{} performs comparably with the state-of-the-art methods, ranking first on \pubmed{} and second on \cora{} and \citeseer{} in terms of micro-F1 scores. On heterophilic graphs, \model{} outperforms all other methods by a margin of at least $2.4\%$ on 3 out of 4 datasets. These results indicate that \model{} generalizes well on different types of graphs. 
The results with macro-F1 are summarized in \autoref{tab:macro_f1}. Macro-F1 scores have not been reported widely in the literature yet. Here, we report the macro-F1 since the heterophilic graphs have imbalanced class distributions than the homophilic graphs. As the results show, \model{} outperforms all other methods across all heterophilic graphs in macro-F1. The difference between the two approximation methods is not significant. Except for a few cases, the difference is largely attributed to hyperparameters choices. The vanilla \model{} gives more consistent results than the approximations although the difference seems marginal.

\begin{figure}[t!]
\centering
 \includegraphics[width=0.75\linewidth]{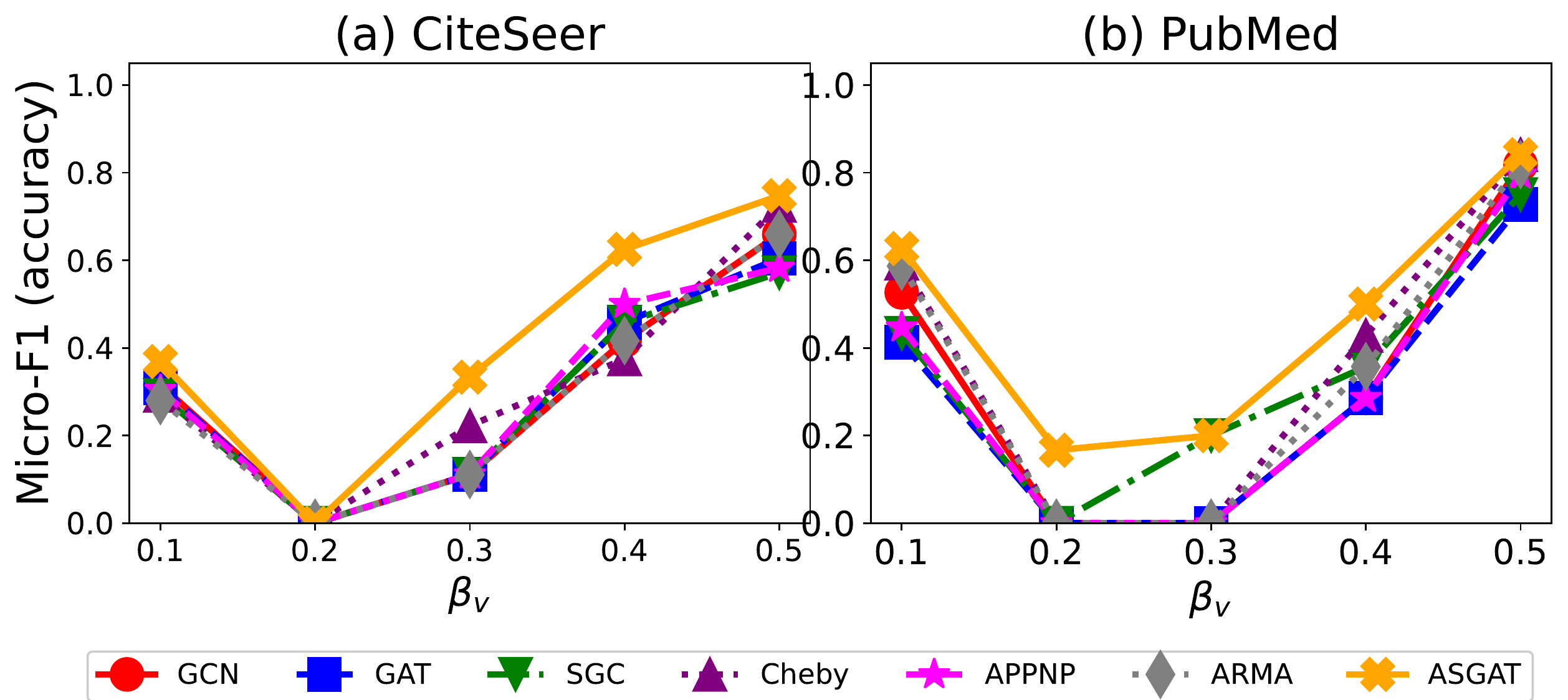}
\hfill
\caption{Micro-F1 results for classification accuracy on heterophilic nodes ($\beta_v\leq 0.5$). \model{} shows better accuracy on classifying heterophilic nodes than the other methods.}
\label{fig:noisy_neighbours}
\end{figure}

Although \model{} performs well on both homophilic and heterophilic graphs, it is unclear how \model{} performs on heterophilic neighbourhoods of an homophilic graph where nodes are mostly of different classes.
Thus, we report an average classification accuracy
on nodes at different levels of $\beta_v$ in \autoref{fig:noisy_neighbours} on the homophilic graphs \citeseer{} and \pubmed{}. The nodes are binned into five groups based on $\beta_v$. For example, all nodes with $0.3 < \beta_v \leq 0.4$ belong to the bin at $0.4$. We have excluded \cora{} from the report since it has very few heterophilic neighbourhoods.

The results in \autoref{fig:noisy_neighbours} show that all models except \model{} perform poorly when $\beta_v$ is low. One may argue that the performance on heterophilic graphs might improve by stacking multiple GNN layers together to obtain information from far-distant nodes. However, it turns out that this approach introduces an oversmoothing problem~\citep{DBLP:conf/aaai/LiHW18} which actually degrades performance. On the other hand, the better performance of \model{} on heterophilic nodes suggests the adaptive spectral filters reduce noise aggregated locally while allowing far-distant nodes to be attended to. 

\begin{figure}[t!]
\centering
 \includegraphics[width=0.8\linewidth]{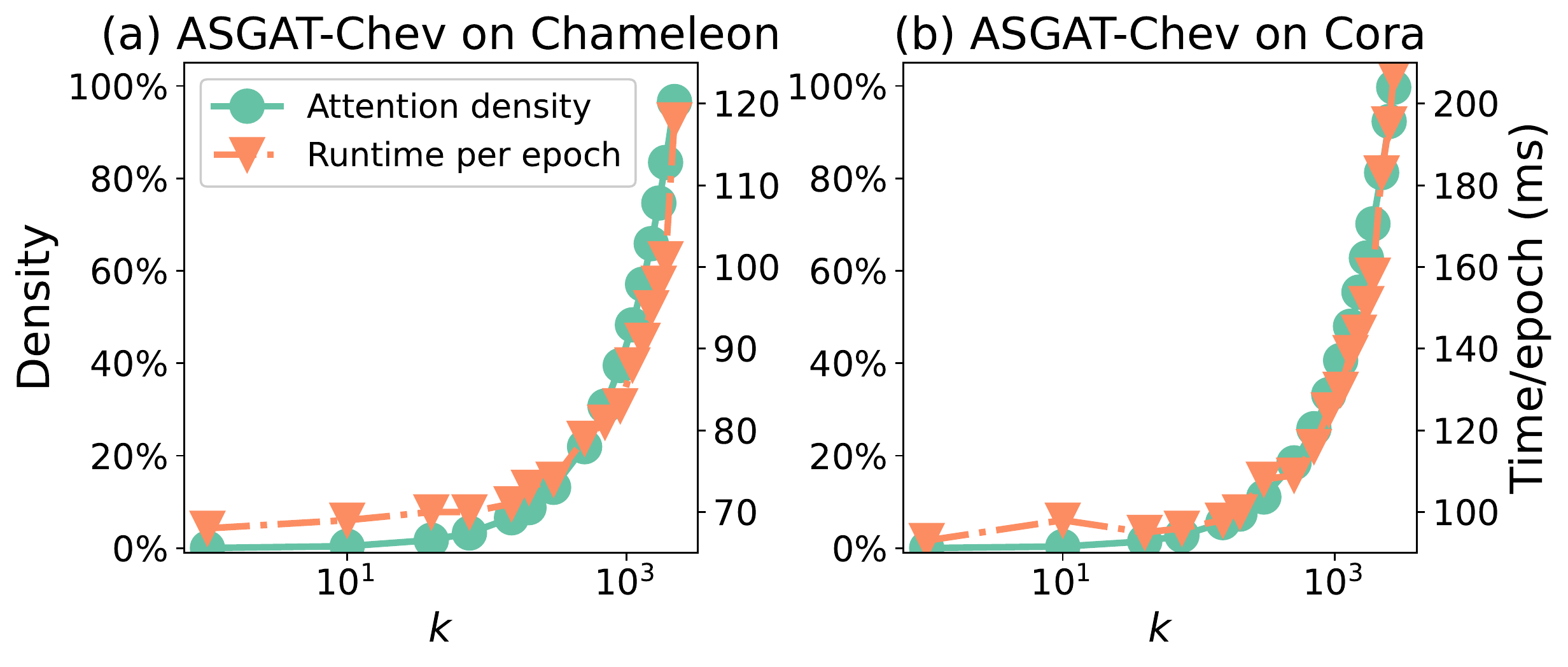}
\hfill
\caption{Attention matrix density and training runtime with respect to $k$. Attention matrix sparsified by keeping the top $k$ elements at each row, which effectively improves runtime efficiency.}
\label{fig:density}
\end{figure}

\textbf{Attention sparsification.}
The restriction on top $k$ entries in \autoref{eqn:psi_k} guarantees a certain level of sparsification. Nonetheless, \model{} requires a partial sorting which adds an overhead of $O(n+k\log N)$. To further analyze the impact of attention sparsity on runtime, we plot the density of an attention matrix with varying $k$ in \autoref{fig:density} along with its running time. The results are drawn from two datasets: the heterophilic dataset \chameleon{} and the homophilic dataset \cora. As expected, \model{} shows a stable growth in the attention density as the value of $k$ increases. It also shows that \model{} runs much faster when attention weights are well-sparsified. In our experiments, we find the best results are achieved on $k<20$.

\begin{figure*}[!t]
\centering
 \includegraphics[width=\textwidth]{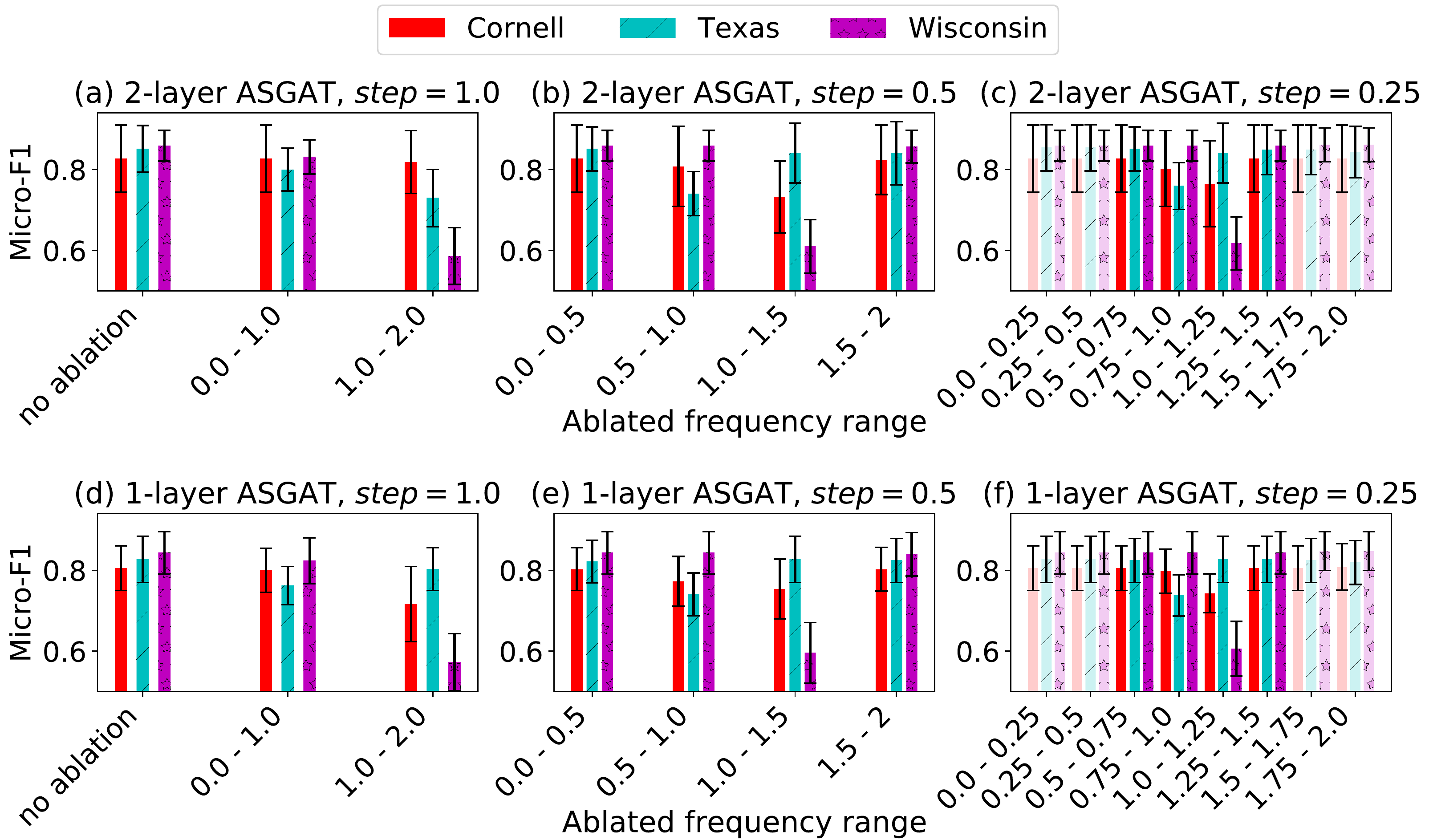}
\hfill
\caption{Micro-F1 with respect to ablated frequency sub-ranges on heterophilic graphs. We divide the frequency range into a set of sub-ranges with different lengths. The results (a) and (d) reveal the importance of high-frequency range $(1\sim2)$. Further experiments show that there is a subtle difference in the most important range across datasets, but it ranges between $(0.75\sim 1.25)$.}
\label{fig:freqency_ablation}
\end{figure*}

\textbf{Frequency range ablation.} To understand how adaptive spectral filters contribute to \model{}'s performance on heterophilic graphs, we conduct an ablation study on spectral frequency ranges. We first divide the entire frequency range $(0\sim2)$ into a set of predefined sub-ranges exclusively, and then manually set the filter frequency responses to zero for each sub-range at a time in order to check the impact of each sub-range on the performance of classification. By doing so, the frequencies within a selected sub-range do not contribute to neither node attention nor feature aggregation, therefore helping to reveal the importance of the sub-range. We consider three different lengths of sub-ranges, i.e., step=$1.0$, step=$0.5$, and step=$0.25$. The results of frequency ablation on the three homophilic graphs are summarized in \autoref{fig:freqency_ablation}.

The results for step=$1.0$ reveal the importance of high-frequency range $(1\sim2)$ on node classification of heterophilic graphs. The performance is significantly dropped by ablating high-frequency range on all datasets. Further investigation at the finer-level sub-ranges (step=$0.5$) shows that ablating sub-range $0.5\sim1.5$ has the most negative impact on performance, whereas the most important sub-range varies across different datasets at the finest level (step=$0.25$).
This finding matches our intuition that low-pass filters used in GNNs underlie a homophily assumption in a similar way as naive local aggregation. We suspect the choice of low-pass filters also relates to oversmoothing issues in spectral methods~\citep{DBLP:conf/aaai/LiHW18}, but we leave it for future work.

\textbf{Attention head ablation.} In \model{}, each head uses a spectral filter to produce attention weights. To delve the importance of a spectral filter, we further follow the ablation method used by \citet{DBLP:conf/nips/MichelLN19}. Specifically, we ablate one or more filters by manually setting their attention weights to zeros. We then measure the impact on performance using micro-F1. If the ablation results in a large decrease in performance, the ablated filters are considered important. We observe that all attention heads (spectral filters) in \model{} are of similar importance, and only all attention heads combined produce the best performance. Please check \autoref{sec:ablation} for the detailed results.

\textbf{Time complexity.} In vanilla \model{}, eigen-decomposition is required for \autoref{eqn:graph_wavelet_matrix} which has a time complexity of $O(N^3)$. \model{}-Cheb and \model{}-ARMA avoid eigen-decomposition and are able to scale to large graphs as their time complexities are $O(R\times|E|)$ and $O((P\times T + Q)\times |E|)$ respectively, where $R$, $P$ and $Q$ are polynomial orders that are normally less than $30$, $T$ is the number of iterations that is normally less than $50$. Therefore, both \model{}-Cheb and \model{}-ARMA scale linearly with the number of edges $|E|$. Readers can refer to \autoref{sec:approximation} for a more detailed introduction of these two methods. Secondly, partial sorting used in the attention sparsification of \autoref{eqn:psi_k} requires $O(N+k \log N)$. Lastly, \autoref{eqn:softmax} is performed on a length-$k$ vector for $N$ rows; therefore, a time complexity of $O(k\times N)$ is needed. In practice, we have $R \sim P \sim T \sim Q \sim k \ll N \ll |E|$ for most graphs, therefore, for a model with $M$ heads, the overall time complexity is $O(M\times R\times|E|)$ for \model{}-Cheb and $O(M \times (P\times T + Q)\times |E|)$ for \model{}-ARMA.

\section{Related Work}
\label{related_work}


Graph neural networks have been extensively studied recently. We categorize work relevant to ours into three perspectives and summarize the key ideas.

\textbf{Attention on graphs.} 
Graph attention networks (GAT) \citep{DBLP:conf/iclr/VelickovicCCRLB18} was the first to introduce attention mechanisms on graphs. GAT assigns different importance scores to local neighbors via an attention mechanism. Similar to other GNN variants, long-distance information propagation in GAT is realized by stacking multiple layers together. Therefore, GAT suffers from the oversmoothing issue \citep{DBLP:conf/iclr/ZhaoA20}. 
\citet{DBLP:conf/iclr/0001ZWZ20} improve GAT by incorporating both structural and feature similarities while computing attention scores.

\textbf{Spectral graph filters and wavelets.} Some GNNs also use graph wavelets to extract information from graphs. \citet{DBLP:conf/iclr/XuSCQC19} applied graph wavelet transform defined by \citet{DBLP:journals/spm/ShumanNFOV13} to GNNs. \citet{DBLP:conf/nips/KlicperaWG19} proposed a general GNN argumentation using graph diffusion kernels to rewire the nodes. \citet{DBLP:conf/kdd/DonnatZHL18} used heat wavelet to learn node embeddings in unsupervised ways and showed that the learned embeddings closely capture structural similarities between nodes. Other spectral filters used in GNNs can also be viewed as special forms of graph wavelets~\citep{DBLP:conf/iclr/KipfW17, DBLP:conf/nips/DefferrardBV16, DBLP:journals/corr/abs-1901-01343}. Coincidentally, \citet{DBLP:journals/corr/abs-2003-07450} also noticed useful information carried by high-frequency components from a graph Laplacian. Similarly, they attempted to utilize such components using node attentions. However, they resorted to the traditional choice of heat kernels and applied such kernels separately to low-frequency and high-frequency components divided by a hyperparameter. In addition to this, their work did not link high-frequency components to heterophilic graphs.

\textbf{Prediction on heterophilic graphs.} \citet{Pei2020} have drawn attention to GCN and GAT's poor performance on heterophilic graphs very recently. They try to address the issue by essentially pivoting feature aggregation to structural neighborhoods from a continuous latent space learned by unsupervised methods. Another attempt to address the issue was proposed by \citet{Liu2020}. They proposed to sort locally aggregated node embeddings along a one-dimensional space and used a one-dimensional convolution layer to aggregate embeddings a second time. By doing so, non-local but similar nodes can be attended to. Very recently, \citet{DBLP:conf/nips/ZhuYZHAK20} showed a heuristic combination of ego-, neighbor and higher-order embedding improves GNN performance on heterophilic graphs. Coincidentally, they also briefly mentioned the importance of higher-frequency components on heterophilic graphs, but they did not provide an empirical analysis.

Although our method shares some similarities in motivation with the aforementioned work, it is fundamentally different in several aspects. To the best of our knowledge, our method is the first architecture we know that computes multi-headed node attention weights purely from learned spectral filters. As a result, in contrast to commonly used heat kernel, our method utilizes higher-frequency components of a graph, which helps prediction on heterophilic graphs and neighbourhoods.

m\section{Conclusion}
In this paper, we study the node classification tasks on graphs where local homophily is weak. We argue the assumption of homophily is the cause of poor performance on heterophilic graphs. In order to design more generalizable GNNs, we suggest that a more flexible and adaptive feature aggregation scheme is needed. To demonstrate, we have introduced the adaptive spectral graph attention network (\model{}) which achieves flexible feature aggregation using learnable spectral graph filters. By utilizing the full graph spectrum adaptively via the learned filters, \model{} is able to aggregate features from nodes that are close and far. For node classification tasks, \model{} outperforms all benchmarks on heterophilic graphs, and performs comparably on homophilic graphs. On homophilic graphs, \model{} also performs better for nodes with weak local homophily. Through our analysis, we find the performance gain is closely linked to the higher end of the frequency spectrum.

\section*{Acknowledgement}
This work was partly supported by the National Research Foundation of Korea(NRF) grant funded by the Korea government(MSIT) (No. 2020R1F1A1061667).

\appendix
\clearpage

\section*{Appendix}

\section{Further Experiment Results \& Hyperparameter Details}
\label{sec:macro-f1}

\textbf{Macro-F1.} We provide the macro-F1 scores on the classification task in \autoref{tab:macro_f1}. The proposed model outperforms the other models on heterophilic graphs and performs comparable on the homophilic graphs.
\begin{table*}[h]
        \caption{Macro-F1 for node classification task. The proposed model consistently outperforms other methods on heterophilic graphs and performs comparably on homophilic graphs.}
        \label{tab:macro_f1}
        \centering
        \scalebox{0.77}{
         \begin{tabular}{l | r  r  r | r  r  r  r  r  r  r  r  r  r} 
             \toprule
              & \multicolumn{8}{c}{$\text{Homophily} \xLeftrightarrow{\hspace*{6cm}} \text{Heterophily}$} \\
              & \multicolumn{1}{c}{\cora} & \multicolumn{1}{c}{\pubmed} & \multicolumn{1}{c|}{\citeseer} & \multicolumn{1}{c}{\chameleon} &
              \multicolumn{1}{c}{\squirrel} & \multicolumn{1}{c}{\wisconsin} & \multicolumn{1}{c}{\cornell} & \multicolumn{1}{c}{\texas} \\
             \midrule
             GCN & $86.0\pm 0.3$ & $86.8\pm0.2$ & $72.0\pm1.6$ & $-$ & $-$ & $37.6\pm9.2$ & $24.1\pm9.1$& $34.0\pm5.7$ \\
             ChevNet & $86.8\pm0.3$ & $88.7\pm0.3$ & $74.1\pm1.0$ & $65.9\pm2.4$ &  $38.6\pm3.2$ & $52.9\pm7.7$ & $53.6\pm17.6$ & $64.1\pm12.4$ \\
             ARMANet & $80.2\pm6.8$ & $81.6\pm13.9$ & $66.4\pm0.4$ & $60.7\pm6.3$&  $47.3\pm3.7$ & $53.3\pm7.1$ & $48.5\pm9.3$ & $69.1\pm12.3$ \\
             GAT & $86.4\pm0.4$ & $81.6\pm0.1$ & $69.2\pm1.0$ & $-$ &  $-$ & $30.0\pm5.2$ & $19.0\pm2.8$ & $26.5\pm6.8$ \\
             SGC & $86.0\pm0.3$ & $79.8\pm0.3$ & $74.7\pm1.2$ & $31.3\pm4.4$ &  $45.3\pm1.8$ & $35.9\pm6.3$ & $21.9\pm8.5$ & $23.2\pm7.5$ \\
             GraphSAGE & $85.2\pm 0.1$ & $88.7\pm0.6$ & $74.2\pm 0.6$ & $51.6\pm0.4$ & $-$ & $63.7\pm12.4$ & $49.2\pm10.9$ & $62.9\pm9.6$ \\
             APPNP & $\textbf{87.0}\pm0.4$ & $84.8\pm0.3$ & $70.2\pm1.4$ & $44.0\pm1.8$ &  $29.0\pm2.0$ & $55.8\pm5.7$ & $39.6\pm16.6$ & $61.0\pm8.8$ \\
             Geom-GCN & $85.1\pm0.3$ & $88.5\pm0.1$ & $76.9\pm0.5$ & $-$ & $-$ &$-$ & $-$ & $-$ \\
             $\text{H}_2\text{GCN}$ & $86.7\pm0.4$ & $88.2\pm0.4$ & $72.1\pm1.9$ & $59.1\pm1.8$& $37.5\pm3.2$ & $66.9\pm11.8$ & $57.3\pm12.4$ & $59.0\pm6.6$\\
             \midrule
             MLP & $67.2\pm2.5$ & $88.1\pm0.2$ & $67.6\pm3.5$ & $44.3\pm3.0$ &  $26.6\pm2.5$ & $54.6\pm11.5$ & $63.0\pm12.3$ & $61.7\pm15.2$ \\
             \midrule
             Vanilla \model{} & $-$ & $-$ & $-$ & $-$ & $-$ & $\textbf{67.0}\pm10.0$ & $67.7\pm8.8$ & $66.1\pm8.0$ \\
             \model{}-Cheb & $86.9\pm0.3$ & $\textbf{89.2}\pm1.0$ & $\textbf{79.3}\pm0.6$ & $\textbf{68.5}\pm2.1$ & $\textbf{55.8}\pm3.3$ & $64.6\pm6.2$ & $66.6\pm15.3$ & $\textbf{72.0}\pm10.6$ \\
             \model-ARMA & $86.0\pm1.2$ & $87.9\pm1.3$ & $75.1\pm2.0$ & $65.8\pm2.2$ & $50.9\pm3.3$ & $63.1\pm10.3$ & $\textbf{68.2}\pm10.0$ & $60.8\pm16.3$ \\
             \bottomrule
        \end{tabular}
        }
\end{table*}

\begin{figure*}[ht]
\centering
 \includegraphics[width=0.75\textwidth]{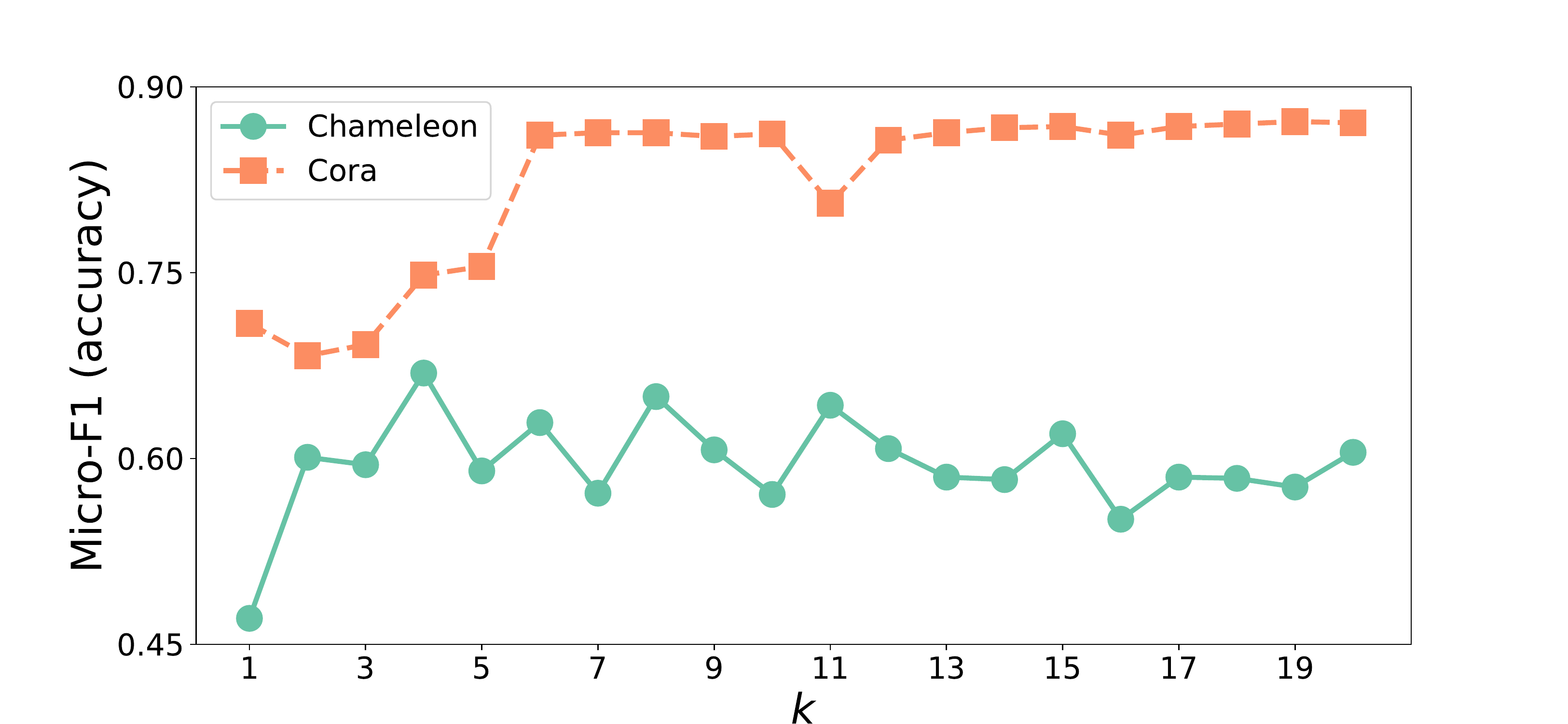}
\hfill
\caption{Micro-F1 score with respect to the sparsification parameter $k$.}
\label{fig:k_performance}
\end{figure*}

\noindent\textbf{Sparsification parameter $k$.} We hereby show how the sparsification parameter $k$ in \autoref{eqn:psi_k} influences node classification peformance on a homophilic graph (\cora{}) and a heterophilic graph (\chameleon{}) in \autoref{fig:k_performance}. The performance on \cora{} increases with $k$ and stabilizes at $6<k<20$, with the exception of $k=11$. In comparison, the performance on \chameleon{} is more sensitive to $k$, as the prediction accuracy fluctuates when $k>1$. The sensitivity to $k$ can be partially explained by the lack of homophilic communities on heteraphic graphs.

\medskip
\noindent\textbf{Hyperparameter details.} The optimal hyperparameters for \model{} are obtained by grid search. For all the benchmark datasets, we set the polynomial order R=15 for \model{}-Cheb, and P=12, Q=18 and T=30 for \model{}-ARMA. The ranges of grid search for other hyperparameters are summarised in \autoref{tab:hyperparameter}.

\begin{table*}[h]
        \caption{The grid search space for the hyperparameters.}
        \label{tab:hyperparameter}
        \centering
         \begin{tabular}{l | c} 
             \toprule
              Hyperparameter & Range \\
             \midrule
             Learning rate & 1e-4, 5e-4, 1e-3, 5e-3, 1e-2, 5e-2 \\
             Hidden size & 32, 64, 128, 256, 512 \\
             Weight decay & 1e-5, 1e-4, 1e-3 \\
             Heads & 2 - 18\\
             Dropout & 0.1, 0.2, 0.4, 0.6, 0.8\\
             k (in \autoref{eqn:psi_k}) & 3 - 18 \\
             
             \bottomrule
        \end{tabular}
\end{table*}

\section{Graph Spectral Filtering Without Eigen-decomposition}
\label{sec:approximation}
Graph filtering is an active research field. Polynomial approximator and rational approximator are the two most well-known classes that are commonly used to approximate eigen-decomposition in graph filtering. While \model{} is agnostic to approximation techniques, the technique used has a slight impact on prediction performance due to errors in approximating \autoref{eqn:graph_wavelet_matrix}. It is also worth noting that, although not discussed in this paper, there exist other approximation methods like Jackson-Chebychev polynomials~\citep{Napoli2016}. Below, we briefly discuss two approaches: Chebyshev and ARMA, which we have used in the experiment.

\subsection{Chebyshev Polynomial Approximation}
\label{subsec:chebyshev}
Chebyshev polynomials approximation is the \textit{de-facto} method for graph Fourier transform and is commonly used in previous works \citep{Hammond2011, Sakiyama2016, DBLP:conf/iclr/XuSCQC19}. In Chebyshev polynomial approximation, a graph signal $\vec{x}$ filtered by a filter $g(\mL)$ is represented as a sum of recursive polynomials~\citep{Sakiyama2016}:
\begin{equation}
\label{eqn:cheby_approx}
g(\mL)\vec{x} = \Big\{\frac{1}{2}c_0 + \sum_{i=1}^p c_i\bar{T}_i(\mL)\Big\}\vec{x}
\end{equation}
where $\bar{T}_0(\mL) = 1$, $\bar{T}_1(\mL)=2(\mL-1)/\lambda_{\max}$, $\bar{T}_i(\mL) = 4(\mL-1)\bar{T}_{i-1}(\mL)/\lambda_{\max} - \bar{T}_{i-2}(\mL)$, and 
\begin{equation}
\begin{aligned}
\label{eqn:cheby_coe}
c_i = \frac{2}{S}\sum_{m=1}^S & \cos\Big(\frac{\pi i (m-\frac{1}{2})}{S}\Big)\cdot \\
& g\Big(\frac{\lambda_{\max}}{2}\Big( \cos\Big(\frac{\pi(m-\frac{1}{2})}{S} + 1\Big)\Big)\Big)
\end{aligned}
\end{equation}
for $i=0,...,R$, where $R$ is the approximation order, $S$ is the number of sampling points and is normally set to $S=R+1$.

In \autoref{eqn:graph_wavelet_matrix}, MLP is used to produce filter responses, i.e., $g=\text{MLP}$. The parameters in MLP are learned by gradient decent from a loss function via the Chebychev approximation. The above approximation has a time complexity of $O(R\times|E|)$. Naturally, approximation error reduces while a larger $R$ is used, which is also why we have $R>12$ in our model. Please note, while Chebyshev polynomials are mentioned in both our method and ChevNet, they are used in fundamentally different ways: ChevNet uses the Chebyshev polynomials with a small $R$ value as a polynomial filter directly, while we refer to its traditional usage in graph signal processing as a method to approximate the eigen-decomposition operation. Since approximation error reduces when a larger $R$ is used, we thus uses $R>12$ in our implementation. 

\subsection{Auto-Regressive Moving-Average (ARMA) Rational Approximation}
As a rational filter, ARMA is known to be more accurate than Chebyshev polynomials~\citep{Isufi2017, Liu2019}. In ARMA, a spectral graph filter $g(L)$ is defined as a rational function:
\begin{equation}
\tilde{g}(\Lambda) = \frac{\sum_{q=0}^Q b_q\Lambda^q}{1 + \sum_{p=1}^Pa_p\Lambda^p}
\end{equation}
where $P$ and $Q$ are the hyper-parameters of polynomial order. $a_p$ and $b_q$ are computed by minimizing
\begin{equation}
\label{eqn:arma_min}
    \min_{a_p,b_q} \Big|\Big|\tilde{g}(\Lambda) - g(\Lambda)\Big|\Big|^2
\end{equation}

\autoref{eqn:arma_min} is solved iteratively with the maximum number of iterations $T$. There are different methods to calculate the filtered graph signals after obtaining $a_p$ and $b_q$, we adopt the conjugate gradient method in the implementation, where the filtered graph signal $\vec{y}$ is obtained by solving the linear system~\citep{Liu2019}

\begin{equation}
    \sum^P_{p=0}a_pL_p\vec{y} = \sum_{q=0}^Qb_qL^q\vec{x}
\end{equation}

This ARMA filter has three hyper-parameters $P$, $Q$ and $T$. Its time complexity is $O((P\times T + Q)\times |E|)$ which linearly scales with $|E|$. ARMA filters are slightly less efficient than Chebyshev.

\section{Ablation study on filters}
\label{sec:ablation}

\begin{figure*}[ht]
\centering
 \includegraphics[width=1\textwidth]{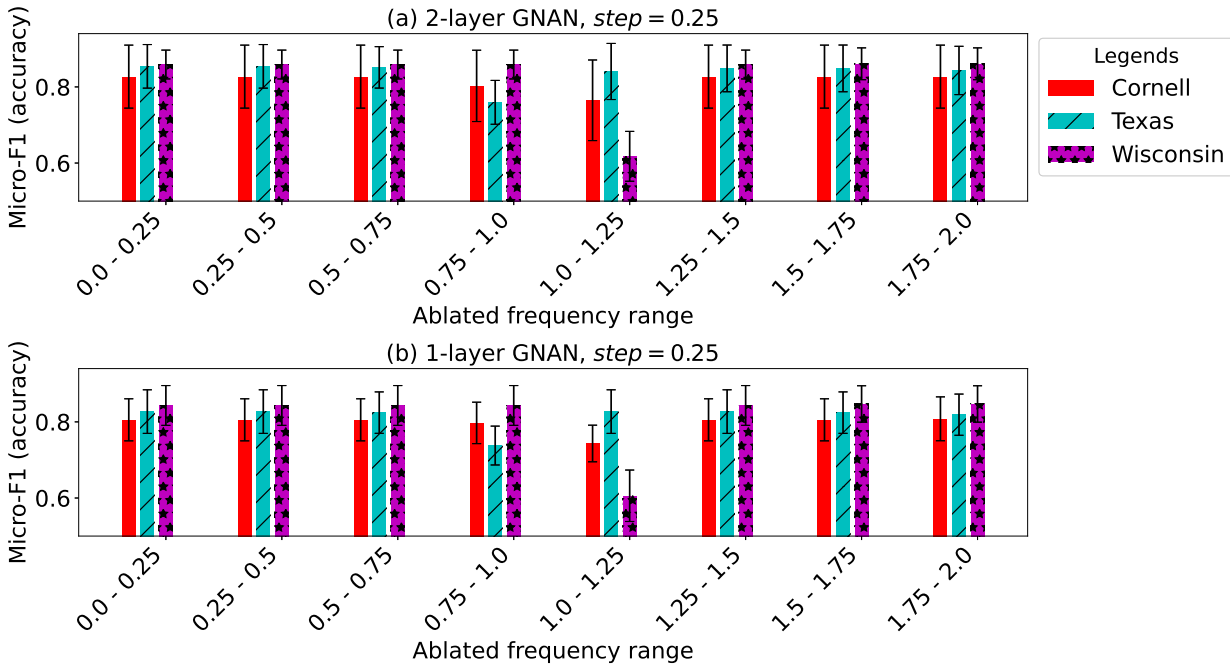}
\hfill
\caption{Full details of the performances on frequency ablation at 0.25 level.}
\label{fig:freqency_ablation_detail}
\end{figure*}

We provide further details for \autoref{fig:freqency_ablation} (c) and \autoref{fig:freqency_ablation} (f) in \autoref{fig:freqency_ablation_detail}.

We further ablated attention heads to check the importance of each head in the prediction tasks. Specifically, we ablate one or more attention heads by manually setting the their attention weights to zeros. We then measure the impact on performance using micro-F1. If the ablation results in a large decrease in performance, the ablated head(s) is considered important. We conduct two types of ablation following the method used by~\citet{DBLP:conf/nips/MichelLN19}.

\medskip
\noindent\textbf{Ablating all but one spectral filter}. To understand how each attention head performs, we keep only one attention head and ablate all others. The results are summarized in \autoref{table:all_but_one_filter_ablation}. We notice that ablating any head results in a decrease in prediction performance. While the accuracy decrease varies per head, the variance is small, indicating all attention head in \model{} are of similar importance, and only all heads combined produces the best performance.

\medskip
\noindent\textbf{Ablating only one attention head}. We then examine performance differences by ablating one head only and keeping all other heads \autoref{table:one_filter_ablation}. Different from the above, ablation of some heads individually only results in a small performance decrease, while performance is intact when other heads are ablated individually. This is an indicator of potential redundancies in the attention heads. We leave the redundancy reduction in the model for future work.

\begin{table*}[ht]
    \caption{Ablation study on attention head. We use 12 attention heads for \cora{} and \pubmed{}, and 14 heads for \citeseer{}.}
    
    \begin{subtable}[ht]{1\textwidth}
        \centering
        \caption{Test accuracy by keeping only one head}     
        \label{table:all_but_one_filter_ablation}
        \scalebox{0.7}{
         \begin{tabular}{c |  c  c  c  c  c  c  c  c  c  c  c  c c c} 
             \toprule
             \\[-.7em]
             \backslashbox{Dataset}{Head} & 1 & 2 & 3 & 4 & 5 & 6 & 7 & 8 & 9 & 10 & 11 & 12 & 13 & 14  \\
             \midrule
             \midrule
             \cora & -2.10\% & -2.10\% & -1.30\% & -2.10\% & -1.50\% & -2.10\% & -2.10\% & -2.10\% & -1.40\% & -1.50\% & -2.10\% & -1.30\% & - & -\\
             \citeseer & -1.8\% & -2.0\% & -1.7\% & -1.7\% & -1.8\% & -1.8\% & -1.7\% & -1.9\% & -1.7\% & -1.9\% & -1.7\% & -1.7\% & -1.9\% & -1.9\%\\
             \pubmed & -4.30\% & -4.40\% & -4.40\% & -5.40\% & -4.30\% & -5.40\% & -5.40\% & -5.40\% & -4.30\% & -4.30\% & -4.30\% & -5.40\% & - & -\\
             \bottomrule
        \end{tabular}
        }
    \end{subtable}
    
    \vspace{1em}
    
    \begin{subtable}[ht]{1\textwidth}
        \centering
        \caption{Test accuracy by ablating only one head}
        \label{table:one_filter_ablation}
        \label{tab:week2}
        \scalebox{0.75}{
         \begin{tabular}{c |  c  c  c  c  c  c  c  c  c  c  c  c c c} 
         \toprule
         \\[-.7em]
         \backslashbox{Dataset}{Head} & 1 & 2 & 3 & 4 & 5 & 6 & 7 & 8 & 9 & 10 & 11 & 12 & 13 & 14  \\
         \midrule
         \midrule
         \cora & 0.00\% & 0.00\% & -0.30\% & 0.00\% & -0.40\% & 0.00\% & 0.00\% & 0.00\% & -0.40\% & -0.40\% & 0.00\% & -0.40\% & - & -\\
         \citeseer & -0.20\% & -0.30\% & -0.70\% & -0.70\% & -0.40\% & -0.60\% & -0.80\% & 0.00\% & -0.60\% & 0.00\% & -0.80\% & -0.50\% & 0.00\% & 0.00\% \\
         \pubmed & -0.80\% & -0.80\% & -0.70\% & 0.00\% & -0.80\% & 0.00\% & 0.00\% & 0.00\% & -0.70\% & -0.80\% & -0.80\% & 0.00\% & - & - \\
         \bottomrule
        \end{tabular}}
     \end{subtable}

     \label{tab:temps}
\end{table*}

\section{Connections to Other Methods}
\label{sec:connection}

In this section, we show \model{} has strong connection to existing models, and many GNNs can be expressed as a special case of \model{} under certain conditions.

\subsection{Connection to GCN}

A GCN~\citep{DBLP:conf/iclr/KipfW17} layer can be expressed as

$$\vh_v^{(l)} = \operatorname{ReLU}(\sum^N_{u=1}\hat{a}_{vu}\vh_u^{(l-1)}\mW^{(l)})$$

where $\hat{a}_{vu}$ is the elements from the $v$-th row of the symmetric adjacency matrix

$$\hat{\mA}=\tilde{\mD}^{-1/2}\tilde{\mA}\tilde{\mD}^{-1/2}  \quad
\text{where} \quad \tilde{A} = \mA + \mI_N,\; \tilde{\emD}_{vv} = \sum_{u=1}^N\tilde{\emA}_{vu}$$

So that

$$
\hat{a}_{vu} = \begin{cases} 1 & \text{if} \; e_{vu} \in E \\ 0 & if \; e_{vu} \notin E\end{cases}
$$

Therefore, GCN can be viewed as a case of \autoref{eqn:layer} with $\sigma = \operatorname{ReLU}$ and $a_{vu} = \hat{a}_{vu}$

\subsection{Connection to Polynomial Filters}

Polynomial filters localize in a node's $K$-hop neighbors utilizing $K$-order polynomials~\citep{DBLP:conf/nips/DefferrardBV16}, most of them takes the following form:

$$g_\theta(\Lambda) = \sum^{K-1}_{k=0}\theta_k\Lambda^k$$

where $\theta_k$ is a learnable polynomial coefficient for each order. Thus a GNN layer using a polynomial filter becomes

$$\vh^{(l)}_v=\operatorname{ReLU}(\sum_{u=1}^N \mU g_\theta(\Lambda)\mU^T\vh_u^(l-1))$$

which can be expressed using \autoref{eqn:layer} with $\mW^{(k)} = \mI_N$, $\sigma = \operatorname{ReLU}$ and $a_{vu}=(\mU g_\theta(\Lambda)\mU^T)_{vu}$. In comparison, our method uses a MLP to learn the spectral filters instead of using a polynomial filter. Also, our method introduces an attention mechanism on top of the filtering response.

\subsection{Connection to GAT}

Our method is inspired by and closely related to GAT~\citep{DBLP:conf/iclr/VelickovicCCRLB18}. To demonstrate the connection, we firstly define a matrix $\Phi$ where each column $\bm{\phi}_v$ is the transformed feature vector of node $v$ concatenated with feature vector of another node (including node $v$ itself) in the graph.
\begin{equation}
    \bm{\phi}_v^{(l)} = ||_{j=0}^N [\mW\vh_v^{(l-1)}||\mW\vh_u^{(l-1)}],
\end{equation}
where $\mW$ is a shared weight matrix, $h_v^{(l-1)}$ and $h_u^{(l-1)}$ are the representation for node $v$ and $u$ from the layer $l-1$. GAT multiplies each column of $\Phi$ with a learnable weight vector $\bm{\alpha}$ and masks the result with the adjacency $\mA$ before feeding it to the nonlinear function $\operatorname{LeakyRelu}$ and $\operatorname{softmax}$ to calculate attention scores. The masking can be expressed as a Hadamard product with the adjacency matrix $\mA$ which is the congruent of a graph wavelet transform with the filter $g(\Lambda) = \mI - \Lambda$:
\begin{equation}
    \Psi = \mA = \mD^{\frac{1}{2}}\mU(\mI - \Lambda)\mU^T \mD^{\frac{1}{2}}
\end{equation}
The GAT attention vector for node $v$ becomes
\begin{equation}
    \va_v = \operatorname{softmax}(\operatorname{LeakyReLU}(\bm{\alpha}^T\bm{\phi}_v \odot \bm{\bar{\psi}}_v))
\end{equation}
where $\bm{\bar{\psi}}_v$ is the $v$-th row of $\Psi$, $\odot$ denotes the Hadamard product, as in \citet{DBLP:conf/iclr/VelickovicCCRLB18}.

In comparison with our method, GAT incorporates node features in the attention score calculation, while node attentions in our methods are purely computed from the graph wavelet transform. Attentions in GAT are restricted to node $v$'s one-hop neighbours only.

\subsection{Connection to Skip-gram methods}

Skip-gram models in natural language processing are shown to be equivalent to a form of matrix factorization \citep{DBLP:conf/nips/LevyG14}. Recently \citet{DBLP:conf/wsdm/QiuDMLWT18} proved that many Skip-Gram Negative Sampling (SGNS) models used in node embedding, including DeepWalk~\citep{DBLP:conf/kdd/PerozziAS14}, LINE~\citep{DBLP:conf/www/TangQWZYM15}, PTE~\citep{DBLP:conf/kdd/TangQM15}, and node2vec~\citep{DBLP:conf/kdd/GroverL16}, are essentially factorizing implicit matrices closely related to the normalized graph Laplacian. The implicit matrices can be presented as graph wavelet transforms on the normalized graph Laplacian. For simplicity, we use DeepWalk, a generalized form of LINE and PTE, as an example. \citet{DBLP:conf/wsdm/QiuDMLWT18} shows DeepWalk effectively factorizes the matrix
\begin{equation}
\label{eqn:SGNS_matrix}
\log\left(\frac{\operatorname{vol}(\gG)}{K}(\sum_{r=1}^K\mP^r)\mD^{-1}\right)-\log(b)
\end{equation}
where $\operatorname{vol}(\gG) =\sum_v \emD_{vv}$ is the sum of node degrees, $\mP=\mD^{-1}\mA$ is the random walk matrix, $K$ is the skip-gram window size (number of hops) and $b$ is the parameter for negative sampling. We know that
$$
\mP = \mI - \mD^{-\frac{1}{2}}\mL\mD^{\frac{1}{2}} = \mD^{-\frac{1}{2}}\mU(\mI - \Lambda)\mU^T\mD^{\frac{1}{2}}.
$$
So \autoref{eqn:SGNS_matrix} can be written using normalized graph Laplacian as:
$$
\log\left(\frac{\operatorname{vol}(\gG)}{K}\mD^{-\frac{1}{2}}\sum_{r=1}^K(\mI-\mL)^r\mD^{\frac{1}{2}}\right) - \log(b).
$$
Or, after eigen-decomposition, as:
\begin{equation}\label{DeepWalkMatrix}
\mM = \log\left(\frac{\operatorname{vol}(\gG)}{Tb}\mD^{-\frac{1}{2}}\mU\sum_{r=1}^K(\mI-\Lambda)^r\mU^T\mD^{\frac{1}{2}}\right),
\end{equation}
where $\mU\sum_{r=1}^T(\mI-\Lambda)^r\mU^T$, denoted as $\bm{\psi}_{sg}$, is a wavelet transform with the filter $g_{sg}(\lambda) = \sum_{r=1}^K(1 - \lambda)^r$. Therefore, DeepWalk can be seen a special case of \model{} by substituting MLP in \autoref{eqn:graph_wavelet_matrix} with $g_{sg}$, and $a_vu$ in \autoref{eqn:layer} with
\begin{align*}
    \bm{a}_{vu} = 
    \begin{cases} 
        {\psi}_{vu} & \quad \text{ if } d_{vu} \leq K \\
        0 & \quad \text{ if } otherwise
    \end{cases}
\end{align*}
where $d_{vu}$ is the shortest distance between node $v$ and $u$. Assigning $\mH = \mW =\mI$, $K=1$ and $\sigma(X)=\log(\frac{\operatorname{vol}(\gG)}{Tb}\mD^{-\frac{1}{2}}\mX\mD^{\frac{1}{2}})$. We have
\begin{equation}
    \vh'_v=\operatorname{FACTORIZE}(\sigma(\bm{a_v}))
\end{equation}
where $\operatorname{FACTORIZE}$ is a matrix factorization operator of choice. \citet{DBLP:conf/wsdm/QiuDMLWT18} uses SVD in a generalized SGNS model, where  the decomposed matrix $\mU'$ and $\bm\Sigma'$ from $\mM=\mU'\bm\Sigma' \mV'$ is used to obtain the node embedding $\vh_v = \mU'\sqrt{\bm\Sigma'}$.

\section{Heat Kernel}

\autoref{tab:heat_macro_f1} shows micro-F1 scores after substituting MLP in \autoref{eqn:graph_wavelet_matrix} with a low-pass heat kernel $g(\Lambda) = e^{-s\Lambda}$, where $s$ is a scaling parameter. Results are reported for each dataset on the best performing $s$. While an explicit heat kernel performs comparably on homophilic graphs (see \autoref{tab:micro_f1}), it fails to perform well on heterophilic graphs due to its incapability to capture high-frequency components. This results confirm the importance of an adaptive wavelet kernel for model generalization to heterophilic graphs. Vanilla \model{}-Heat in this table uses the exact eigen-decomposition thus the results are only reported for datasets where eigen-decomposition is feasible. \model{}-Heat-Chev uses the Chebyshev polynomial approximation described in \autoref{subsec:chebyshev}.

\begin{table}[h!]
        \caption{Micro-F1 for node classification task with a low-pass heat kernel}
        \label{tab:heat_macro_f1}
        \centering
        \scalebox{0.9}{
         \begin{tabular}{l | c | c } 
             \toprule
              & Vanilla \model{}-Heat & \model{}-Heat-Chev \\
             \midrule
             \cora & $-$ & $85.5\pm0.2$ \\
             \citeseer & $-$ & $77.4\pm0.8$\\
             \midrule
             \chameleon & $-$ & $63.1\pm2.1$\\
             \wisconsin & $54.1\pm5.1$ & $54.7\pm4.6$\\
             \cornell & $58.9\pm3.1$ & $59.2\pm3.2$ \\
             \texas & $58.6\pm6.5$ & $58.6\pm5.8$\\
             \bottomrule
        \end{tabular}
        }
\end{table}


\bibliographystyle{splncs04nat}  

\end{document}